\PassOptionsToPackage{table}{xcolor}

\documentclass[sigconf]{acmart}

\usepackage{graphicx}
\usepackage{booktabs}
\usepackage{float}
\usepackage{microtype}
\usepackage{enumitem}
\setlist[itemize]{leftmargin=*, noitemsep, topsep=2pt}
\setlist[enumerate]{leftmargin=*, noitemsep, topsep=2pt}

\copyrightyear{2026}
\acmYear{2026}
\setcopyright{cc}
\setcctype{by}
\acmConference[KDD '26]{Proceedings of the 32nd ACM SIGKDD Conference on Knowledge Discovery and Data Mining V.2}{August 09--13, 2026}{Jeju Island, Republic of Korea}
\acmBooktitle{Proceedings of the 32nd ACM SIGKDD Conference on Knowledge Discovery and Data Mining V.2 (KDD '26), August 09--13, 2026, Jeju Island, Republic of Korea}
\acmDOI{10.1145/3770855.3818913}
\acmISBN{979-8-4007-2259-2/2026/08}

\setlist{noitemsep,topsep=2pt,parsep=0pt,partopsep=0pt}

\begin{document}

\setlength{\textfloatsep}{5pt plus 1.0pt minus 2.0pt}
\setlength{\floatsep}{5pt plus 1.0pt minus 2.0pt}
\setlength{\intextsep}{5pt plus 1.0pt minus 2.0pt}

\title[Machine Learning for Depression Screening and Intervention: an Original Circadian Rhythm \\ Score-based Methodology]{Machine Learning for Depression Screening and Intervention: an Original Circadian Rhythm Score-based Methodology}

\author{Bin Wang}
\authornote{Bin Wang and Shuo Lian are co-first authors with equal contributions.}
\affiliation{%
  \department{Faculty of Information Science and Engineering}
  \institution{Ocean University of China}
  \city{Qingdao}
  \country{China}
}
\email{wangbin9545@ouc.edu.cn}

\author{Shuo Lian}
\authornotemark[1]
\affiliation{%
  \department{Faculty of Information Science and Engineering}
  \institution{Ocean University of China}
  \city{Qingdao}
  \country{China}
}
\email{lianshuo@stu.ouc.edu.cn}

\author{Yuanyuan Hou}

\affiliation{%
  \department{Department of Rehabilitation Medicine}
  \institution{The Affiliated Hospital of Qingdao University}
  \city{Qingdao}
  \country{China}
}
\email{houyuanyuanv@163.com}

\author{Dexian Wang}
\affiliation{%
\department{School of Intelligent Medicine}
  \institution{Chengdu University of Traditional Chinese Medicine}
  \city{Chengdu}
  \country{China}
}
\email{wangdexian@cdutcm.edu.cn}

\author{Peilan He}
\affiliation{%
  \department{Faculty of Information Science and Engineering}
  \institution{Ocean University of China}
  \city{Qingdao}
  \country{China}
}
\email{hepeilan@ouc.edu.cn}

\author{Feng Hong}
\authornote{Feng Hong and Yanwei Yu are corresponding authors.}
\affiliation{%
  \department{Faculty of Information Science and Engineering}
  \institution{Ocean University of China}
  \city{Qingdao}
  \country{China}
}
\email{hongfeng@ouc.edu.cn}

\author{Yanwei Yu}
\authornotemark[2]
\affiliation{%
  \department{Faculty of Information Science and Engineering}
  \institution{Ocean University of China}
  \city{Qingdao}
  \country{China}
}
\email{yuyanwei@ouc.edu.cn}

\author{Tianrui Li}
\affiliation{%
\department{School of Computing and Artificial Intelligence}
  \institution{Southwest Jiaotong University}
  \city{Chengdu}
  \country{China}
}
\email{trli@swjtu.edu.cn}

\renewcommand{\shortauthors}{Bin Wang et al.}
\begin{abstract}
Depression screening from large-scale behavioral data is challenged by fragmented circadian indicators, limited interpretability, and the lack of intervention-oriented analysis. Existing approaches typically analyze sleep, activity, and social behaviors in isolation, failing to capture their joint circadian structure. To address this limitation, we first propose the Circadian Rhythm Score (CRS), a composite index that compresses multi-domain daily behaviors into a unified representation of circadian rhythm. CRS is constructed to maximize discriminative power for depression screening while preserving behavioral semantics through non-negativity constraints. Empirical results demonstrate near-lossless compression, where a single CRS retains almost the full predictive capability compared with multiple raw behavioral indicators. Building upon CRS, we develop an interpretable depression screening framework based on gradient-boosted trees and SHAP analysis, revealing nonlinear and saturation-like associations between circadian rhythm and depression risk. Beyond risk prediction, we further integrate interaction modeling and counterfactual regression to estimate heterogeneous and dose-dependent behavioral effects, enabling intervention-oriented reasoning under different circadian contexts. Experiments on the China Health and Retirement Longitudinal Study (CHARLS, n=15,233), demonstrate robust screening performance (ROC-AUC=0.825) and identify actionable behavioral thresholds, including a minimum effective exercise dose of approximately 300 MET-min/week and an optimal restorative nap duration of approximately 65 minutes for sleep-deprived individuals. By bridging supervised representation learning and interpretable modeling, this work provides a scalable framework for depression screening and intervention-aware healthcare data mining.

\end{abstract}



\maketitle

\section{Introduction}

\textbf{Challenges:}
Depression is a prevalent mental disorder characterized by persistent low mood and constitutes a major and growing global health burden, particularly among older adults \cite{steffens2024treatment}. The 2019 Global Burden of Disease study ranked depression as the second leading cause of years lived with disability worldwide \cite{global2022burden}. Depression is strongly associated with reduced quality of life, functional impairment \cite{yan2024association}, increased mortality \cite{patel2015addressing}, and elevated risk of suicidal behavior \cite{coccaro2019new}. In the context of rapid population aging, depression is projected to become the leading contributor to global disease burden by 2030 \cite{hock2012new}. Despite extensive clinical efforts, conventional treatments—primarily pharmacotherapy and psychotherapy—often produce modest effect sizes, suggesting a ceiling in therapeutic efficacy \cite{leichsenring2022efficacy}. Adverse medication effects and poor treatment adherence have further motivated interest in alternative and complementary interventions \cite{garfield2016relationship}. These therapeutic challenges are compounded by systemic barriers, including low rates of early detection and limited access to evidence-based mental healthcare \cite{global2024incidence}. \textit{Consequently, the early screening, and effective management of late-life depression remain urgent priorities for global public health} \cite{herrman2022time}.

\begin{figure*}[t]
  \centering
  \includegraphics[width=\textwidth]{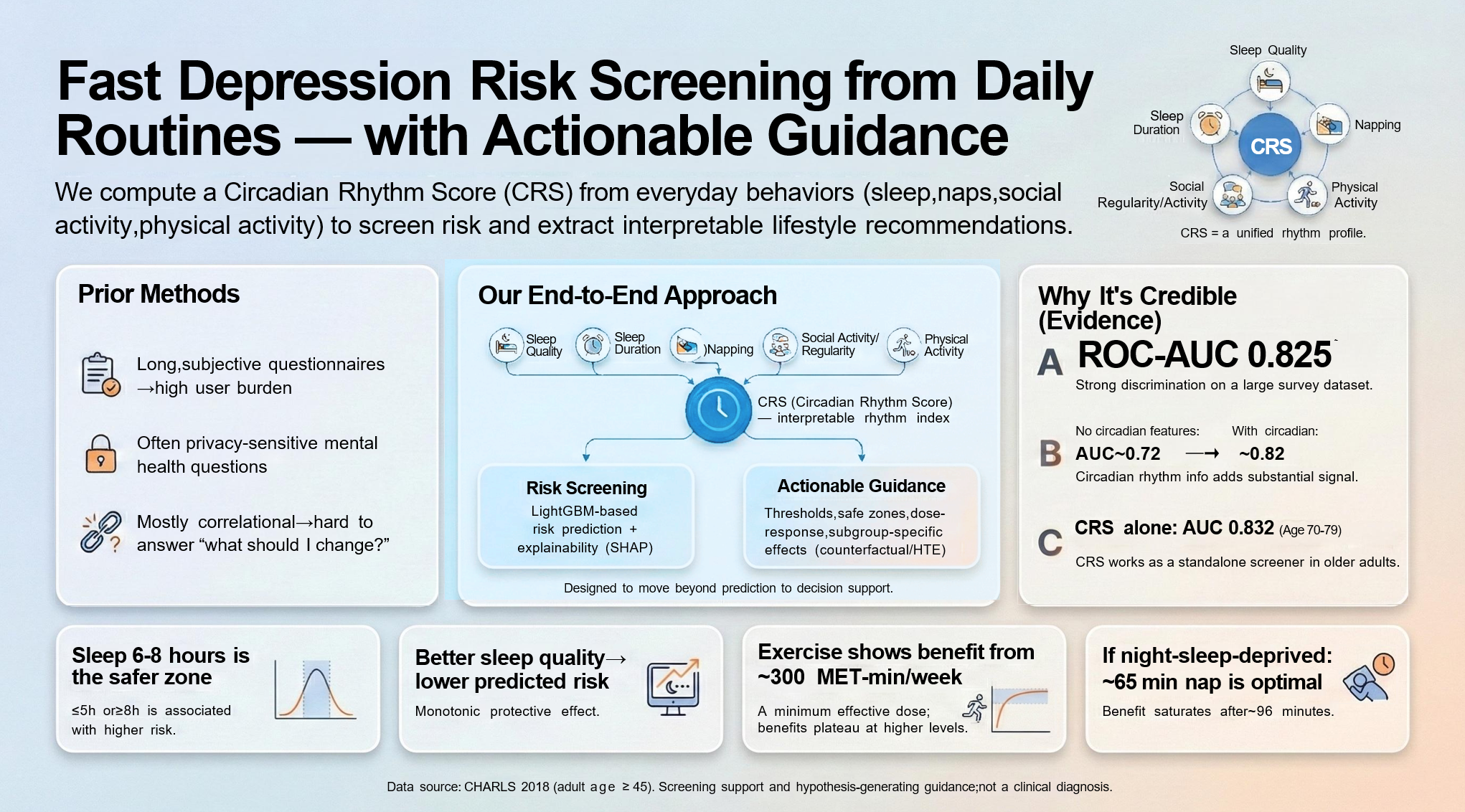}
  \caption{Visual summary of the proposed framework. The framework integrates daily behavioral data into a composite index CRS to achieve effective depression screening (ROC-AUC 0.825) while providing interpretable, dose-dependent lifestyle guidance on sleep, exercise, and napping.}
  
  \Description{A comprehensive infographic titled 'Fast Depression Risk Screening from Daily Routines'. It contrasts 'Prior Methods' (subjective questionnaires) with the proposed 'End-to-End Approach' (CRS-based screening and actionable guidance). Key results highlight an ROC-AUC of 0.825. The bottom panels illustrate specific intervention insights: a protective sleep window of 6-8 hours, a minimum effective exercise dose of 300 MET-min/week, and an optimal nap duration of 65 minutes for sleep deprived individuals.}
  \label{fig:visual_summary}
\end{figure*}

\textbf{Motivation:} Depression is characterized by impaired neuroplasticity and structural abnormalities in key brain regions, including the hippocampus and prefrontal cortex \cite{pittenger2007stress, koolschijn2009brain, schmaal2015subcortical}. Growing evidence indicates that circadian rhythm disruption—reflecting misalignment of the endogenous biological clock—is not merely a secondary manifestation but a core mechanism contributing to depressive pathogenesis \cite{crouse2021circadian, shen2023circadian}, a view supported by \textit{\textbf{our prior studies in animal models demonstrating that circadian misalignment impairs hippocampal and prefrontal function and induces depressive-like behaviors \cite{hou2021long, hou2025long, zuo2024circadian}.}} Sleep and circadian rhythms are tightly coupled \cite{franken2023sleep}, and chronic sleep disturbances show a bidirectional, self-reinforcing relationship with depression \cite{baglioni2011insomnia, baglioni2016sleep, chaput2022role}, with consistent associations reported for sleep duration and regularity \cite{phillips2017irregular, zhai2015sleep}. This interplay extends to daytime napping, which exhibits heterogeneous and often bidirectional associations with depressive symptoms, particularly in older adults \cite{jeon2023circadian, qin2025bidirectional}. Physical activity further modulates circadian function and mood regulation \cite{noone2024understanding}, with exercise demonstrating antidepressant effects and partial restoration of neuroplasticity in major depressive disorder \cite{bruchle2021physical, vancampfort2025efficacy}. Collectively, sleep, napping, and physical activity shape the circadian rest–activity rhythm (RAR), a behavioral manifestation of circadian health that is strongly associated with depression risk \cite{jeon2023circadian, wang2015suprachiasmatic, kang2024integrative}. Moreover, prior research suggests that jointly addressing cross-domain circadian behaviors may outperform isolated interventions \cite{manber2008cognitive}, while large-scale survey data are typically self-reported and heterogeneous \cite{he2025cross}, underscoring the need for a unified circadian representation. \textit{Motivated by this evidence, we  develop a machine-learning–oriented CRS that integrates sleep, napping, social activity, and physical activity into a unified representation for depression screening and intervention analysis.}

\textbf{Research Gaps:} Although machine learning has been increasingly applied to depression research \cite{durstewitz2019deep}, several critical gaps remain:
\begin{itemize}

  \item \textit{Lack of an integrated representation for circadian rhythm quantification.}
Single-domain features are insufficient to capture systemic circadian risk. Existing studies typically examine sleep or physical activity in isolation, lacking an integrated score that quantifies cross-domain circadian rhythm. This limitation is notable given that circadian regulation spans sleep, social engagement, and physical activity \cite{ehlers1988social, germain2008circadian, walker2020circadian}, all of which have been independently linked to depressive symptoms across populations \cite{au2017relationship, levandovski2011depression, phillips2017irregular}.

 \item \textit{Limited interpretability and actionable insight.}
While modern machine learning models improve predictive accuracy, they often operate as black boxes. For clinical decision support, it is essential not only to identify influential risk factors (e.g., insufficient sleep) but also to characterize interpretable and actionable thresholds that can guide behavioral recommendations.

\item \textit{Emphasis on correlation over intervention.}
Most existing machine learning models focus on learning input–output associations and are therefore unable to address counterfactual or intervention-oriented questions (e.g., How much would depression risk decrease with increased exercise?). As a result, these models struggle to transition from risk prediction tools to decision support systems capable of informing personalized intervention strategies.
\end{itemize}

\textbf{Our Contributions:} To address the above research gaps, we design an analytical framework centered on CRS that enables interpretable, predictive, and intervention-oriented depression screening. The proposed framework identifies CRS as the most influential predictor for depression discrimination and demonstrates that CRS alone achieves strong screening performance, attaining an AUC of up to 0.83 in the 70–79 age group. Beyond risk prediction, the framework integrates causal inference to derive actionable intervention insights, some interesting conclusions like a protective sleep duration window of $\sim$6 hours and a supported minimum effective exercise dose of approximately 300 MET-min/week. 

This work makes the summarized key contributions:

\begin{itemize}
\item We propose the CRS, a supervised and interpretable composite index that integrates heterogeneous daily behaviors into a compact, task-aligned representation of circadian robustness. Unlike prior single-domain or heuristic measures, CRS is constructed by  optimizing screening performance while preserving behavioral semantics.

\item  We demonstrate that CRS achieves near-lossless compression of rich circadian-related features, retaining most discriminative power of the full behavioral set and exhibiting strong standalone screening performance across demographic subgroups. This enables simplified yet effective depression risk assessment at population scale.

\item Building upon CRS-based screening, we integrate interpretable machine learning and counterfactual regression model to analyze heterogeneous and dose-dependent associations of modifiable behaviors under different circadian contexts. This framework supports principled exploration of \textit{who benefits, under what conditions, and within what dosage ranges,} bridging depression screening with actionable, testable intervention hypotheses.

    



\end{itemize}

\section{Related Works}

\textbf{Depression and Circadian Rhythms.} 
Depression is a leading cause of global disability \cite{global2022burden}, with growing evidence implicating circadian rhythm disruption as a core pathophysiological mechanism rather than merely a secondary symptom \cite{walker2020circadian}. 
Disturbances in the sleep-wake cycle, such as insomnia, have been identified as crucial predictors of depression onset \cite{baglioni2011insomnia}. 
Beyond sleep duration, daytime napping exhibits complex, bidirectional associations with depressive symptoms, particularly in older adults \cite{li2022daytime}. 
Similarly, physical activity serves as a critical regulator of biological rhythms, with meta-analyses confirming its antidepressant effects \cite{schuch2018physical}. 
While prior work often examines these behaviors in isolation \cite{phillips2017irregular}, recent approaches advocate for integrated RAR markers to capture systemic circadian health \cite{jeon2023circadian}. 
However, traditional RAR measures often require longitudinal actigraphy, limiting their applicability in large-scale epidemiological screenings \cite{kang2024integrative}.

\textbf{Machine Learning and Causal Intervention.} 
To translate circadian-behavioral insights into scalable population screening, machine learning models have been increasingly applied to large-scale health surveys such as CHARLS \cite{wang2025plosml}. 
Tree-based methods often outperform traditional linear models on tabular health data due to their ability to capture complex nonlinear relationships \cite{grinsztajn2022tree}. 
Despite their predictive success, many existing approaches remain largely black-box, limiting their utility for clinical interpretation and decision support \cite{li2024interpretablecharls}. 
More importantly, prediction alone do not reveal whether modifying a behavior would lead to improved mental health outcomes \cite{herrman2022time}. 
To bridge this gap, To bridge this gap, recent studies have incorporated Shapley-value explanations for model interpretability \cite{lundberg2017unified} and heterogeneous treatment effect estimation for counterfactual reasoning \cite{kunzel2019metalearners}. Nevertheless, existing frameworks rarely combine interpretable causal analysis with a unified circadian indicator, limiting their ability to generate actionable and personalized recommendations for depression prevention.

\section{Methodology}
\begin{figure*}[htb!]
    \centering
    \includegraphics[width=0.95\textwidth]{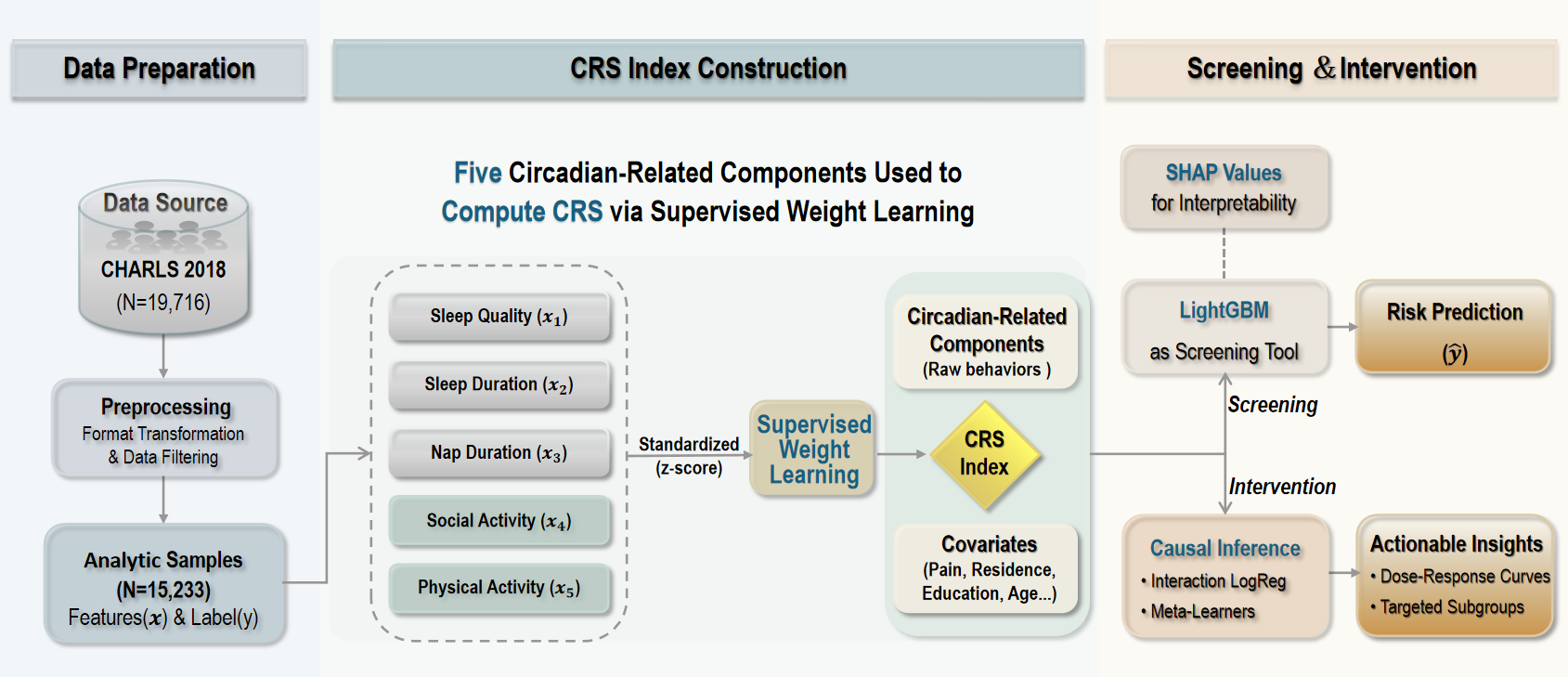}
    \Description{A horizontal flow diagram illustrating the analytical framework. 
    Left: Data Preparation filters CHARLS 2018 data (N=19,716) to analytic samples (N=15,233). 
    Middle: CRS Index Construction. Five behavioral components (Sleep Quality, Sleep Duration, Nap Duration, Social Activity, Physical Activity) are standardized and aggregated via Weight Learning to form the CRS Index (yellow diamond). 
     Right: Screening \& Intervention. The CRS Index is combined with raw circadian behaviors and covariates (e.g., Pain, Residence, Age) to form the feature set. 
    The flow splits into two branches: The 'Screening' branch (top) uses LightGBM for risk prediction, supported by SHAP values. 
    The 'Intervention' branch (bottom) uses Causal Inference (Interaction LogReg, counterfactual inference techniques) to generate Actionable Insights.}
    \caption{\textbf{Overview of the proposed CRS-based analytical framework.} The pipeline comprises three stages: (1) \textbf{CRS Construction}: aggregating five behavioral domains into a unified score using supervised weight learning; (2) \textbf{Screening (Top Branch)}: utilizing the CRS, raw behaviors, and covariates in a LightGBM model for risk prediction, interpreted via SHAP; and (3) \textbf{Intervention (Bottom Branch)}: applying causal inference techniques to extract actionable insights from the data.}
    \label{fig:framework_overview}
\end{figure*}

We propose a circadian rhythm score for depression screening and intervention-oriented analysis, centered on a compact composite indicator termed CRS. The framework consists of three main components: (1) Supervised construction of CRS from multiple behavioral factors; (2) Depression risk screening using a tree-based model with post-hoc interpretability; and (3) Counterfactual analysis to explore dose-dependent behavioral effects for actionable guidance. An overview of the framework is shown in Figure \ref{fig:framework_overview}.


\subsection{Dataset Preparation and Screening Task}
The study population was derived from the 2018 wave of the China Health and Retirement Longitudinal Study (CHARLS) \cite{he2025cross}, a nationally representative survey of middle-aged and older adults in China. Following data preprocessing and standard exclusion procedures, including the removal of participants with missing key variables, the final analytic cohort consisted of 15,233 individuals aged 45 years and above.

\textit{Problem Formulation.} The depression screening task is formulated as a \textit{binary classification problem}. We aim to learn a mapping function $f: X \to \hat{y}$ that predicts the probability of elevated depressive risk. The input component $X$ consists of three aspects:
\begin{equation}\label{ref:xcrs}
    X = \{ \mathbf{x}_{\text{CRS}}, \mathbf{x}_{\text{Rhythm}}, \mathbf{x}_{\text{Covariates}} \}
\end{equation}
where $\mathbf{x}_{\text{CRS}}$ is the proposed composite score CRS, $\mathbf{x}_{\text{Rhythm}}$ denotes the raw circadian behaviors (e.g., sleep duration, nap frequency), and $\mathbf{x}_{\text{Covariates}}$ represents socio-demographic and health-related factors. Detailed variable description is displayed in Table ~\ref{app:control_vars}. The screening model is then represented as:
\begin{equation}
    \hat{y} = \operatorname{LightGBM}(X)
\end{equation}

The ground truth label $y$ is derived according to the CES-D-10 scale \cite{radloff1977ces} (see Appendix \ref{app:gt}) , where participants with a total score $\ge 10$ are labeled as having elevated depressive risk (i.e., $y=1$), consistent with established epidemiological cut-offs \cite{andresen1994screening}. 

\begin{table*}[!ht]
  \centering
  \caption{Description of control variables and coding scheme.}
  \label{app:control_vars}
  \small
  \renewcommand{\arraystretch}{0.72}
  \setlength{\tabcolsep}{3pt}
  \begin{tabular}{@{}p{0.24\textwidth}p{0.70\textwidth}@{}}
    \toprule
    \textbf{Variable} & \textbf{Definition} \\
    \midrule

    \multicolumn{2}{l}{\textbf{Circadian rhythm--related variables}} \\
    \quad \texttt{sleep\_quality\_score} ($x_{i,1}$)
      & Ordinal variable: 1 = very poor, 2 = poor, 3 = fair, 4 = good \\
    \quad \texttt{night\_sleep\_duration\_hr} ($x_{i,2}$)
      & Continuous variable (hours per night) \\
    \quad \texttt{nap\_duration\_min} ($x_{i,3}$)
      & Continuous variable (minutes per day) \\
    \quad Primary\_activity\_purpose ($x_{i,4}$)
      & Categorical variable: 0 = no valid activity, 1 = job-related, 2 = entertainment, 3 = exercise, 4 = other \\
    \quad \texttt{weighted\_social\_score}
      & Used for $x_{i,4}$ when available. Activity-based weighted score: high benefit = 3, moderate benefit = 2, low benefit = 1, no benefit = 0; summed across reported activities \\
    \quad \texttt{social\_activity\_count}
      & Used when \texttt{weighted\_social\_score} is unavailable. Count variable: number of reported social activities \\
    \quad Activity\_regularity
      & Continuous variable (days per week), defined as $\max(\text{da052\_1}, \text{da052\_2}, \text{da052\_3})$ \\
    \quad \texttt{total\_activity\_level\_met} ($x_{i,5}$)
      & Continuous variable (MET (Metabolic Equivalent of Task)-hours per week), intensity-weighted sum with weights 8.0 (vigorous), 4.0 (moderate), and 3.3 (light) \\
    \quad Activity\_intensity
      & Categorical variable: 0 = inactive (all da052\_x = 0); 1 = light-dominant (da052\_3 $>$ 0); 2 = moderate-dominant (da052\_2 $>$ 0); 3 = vigorous-dominant (da052\_1 $>$ 0) \\

    \midrule
    \multicolumn{2}{l}{\textbf{Socio-demographic-related variables}} \\
    \quad Gender & 1 = male, 2 = female \\
    \quad Age & Continuous variable (years) \\
    \quad Residence & 1 = urban, 2 = rural \\
    \quad Education
      & Years of schooling coded as: 1 = illiterate; 3 = incomplete primary school; 6 = primary school; 9 = middle school; 12 = high school/vocational school; 15 = junior college/associate degree; 16 = four-year college/bachelor's degree; 19 = master's degree; 22 = doctoral degree (Ph.D.) \\
    \quad Marital\_status
      & 1 = partnered (married/cohabiting), 2 = single (never married/divorced/widowed) \\
    \quad Work\_life\_status\_code
      & 0 = missing; 1 = working (satisfied); 2 = working (dissatisfied); 3 = working (satisfaction unknown); 4 = recently retired ($\le$ 5 years); 5 = mid-term retired (5--10 years); 6 = long-term retired ($>$ 10 years); 7 = retired (duration unknown); 8 = never worked; 9 = unemployed/other \\

    \midrule
    \multicolumn{2}{l}{\textbf{Health behavior-related variables}} \\
    \quad Smoking\_intensity & Continuous variable (cigarettes per day) \\
    \quad Smoking\_status & 0 = never smoker, 1 = former smoker, 2 = current smoker \\
    \quad Total\_alcohol\_g\_per\_month & Continuous variable (grams per month) \\
    \quad Drink\_status & 0 = never drinker, 1 = former drinker, 2 = current drinker \\

    \midrule
    \multicolumn{2}{l}{\textbf{Health status-related variables}} \\
    \quad Chronic\_diseases\textsuperscript{a}
      & Continuous variable (count of diagnosed conditions) \\
    \quad Pain
      & Binary variable: 0 = no pain, 1 = pain reported \\
    \bottomrule
  \end{tabular}

  \begin{minipage}{0.96\textwidth}
    \footnotesize
    \textsuperscript{a} Chronic diseases include hypertension, hyperlipidemia, hyperglycemia, malignant tumor, chronic lung disease, liver disease, heart disease, stroke, kidney disease, stone disease, arthritis, asthma, mental disorders, and memory-related diseases.
  \end{minipage}
\end{table*}

\subsection{CRS Index Construction}
\label{sec:crs_construction}
Circadian regulation manifests through coordinated daily behaviors, including sleep, napping, physical activity, and social engagement. Existing studies typically model these behaviors independently, which fragments circadian information and complicates interpretation. We address this limitation by constructing a single composite index CRS (i.e., the $\mathbf{x}_{\text{CRS}}$ in Formula \ref{ref:xcrs}) that summarizes overall circadian rhythm while remaining interpretable and predictive.

\subsubsection{Definitions of Circadian-related Components.}
For each individual, we extract five raw circadian-related components $\mathbf{x} = \big(x_{i,1}, \ldots, x_{i,5}\big)^{\top}$ and standardize it as $\mathbf{z}$ (detailed in \textit{Appendix \ref{ref:weight}})  reflecting {sleep quality} ($z_{i,\mathrm{SleepQuality}}$), {nocturnal sleep duration} ($z_{i,\mathrm{NightSleep}}$), {daytime napping} ($z_{i,\mathrm{Nap}}$), {social activity} ($z_{i,\mathrm{SocialAct}}$), and {physical activity} ($z_{i,\mathrm{PhyAct}}$). 
Notably, to capture the distinct physiological impacts of nap duration, daytime napping ($z_{i,\mathrm{Nap}}$) is processed via discrete encoding (rewarding short power naps while penalizing excessive duration), whereas other components like night sleep duration are treated as continuous variables. 
The components are aligned such that higher values contribute positively to CRS.

The $\mathbf{x}_{\text{CRS}}^i$ for individual $i$ is computed as the weighted sum of normalized components $\mathbf{z}_i$, where weights are constrained to be non-negative:
\begin{equation}\label{xcrs}
\mathbf{x}_{\text{CRS}}^i = \mathbf{w}^{\top} \mathbf{z}_i, \quad \mathbf{w} \ge 0,
\end{equation}
where $\mathbf{x}_{\text{CRS}}^i$ and $\mathbf{z}_i$ denote the CRS and standardized component vector for individual $i$, respectively, and $\mathbf{w}$ is the shared weight vector achieved via supervised learning.

\subsubsection{Weight Search via Supervised Learning.} 
Rather than assigning subjective or uniform weights, we determine the optimal weight vector $\mathbf{w^*}$ through a supervised learning approach using LightGBM to maximize depression screening performance. Specifically, we conducted a randomized search over $\mathbf{w}$ with 100,000 trials, sampling each component independently from a uniform distribution $\mathcal{U}(0,1)$ subject to the non-negativity constraint $\mathbf{w} \ge 0$.

For each candidate $\mathbf{w}$, we computed the composite score $\mathbf{x}_{\text{CRS}}$ using Eq.~(\ref{xcrs}). To optimize the ROC-AUC via LightGBM,  $-\mathbf{x}_{\text{CRS}}$ is taken as the input with the maximization of the screening score. The best-performing weight vector was then $L_1$-normalized such that $\|\mathbf{w^*}\|_1=1$ and fixed for all subsequent analyses. This procedure is computationally acceptable, requiring approximately 5 minutes. The resulting optimal weights are:
\[
\mathbf{w^*} = (0.6874,\; 0.1203,\; 0.0091,\; 0.1233,\; 0.0599).
\]
Consequently, the  composite score $\mathbf{x}_{\text{CRS}}^i$ for the individual $i$ is defined as:
\vspace{-4pt}
\begin{equation}
\begin{aligned}
\mathbf{x}_{\text{CRS}}^i
&= 0.6874\, z_{i,\mathrm{SleepQuality}}
 + 0.1203\, z_{i,\mathrm{NightSleep}}\\
&\quad + 0.0091\, z_{i,\mathrm{Nap}}
+ 0.1233\, z_{i,\mathrm{SocialAct}} \\
&\quad + 0.0599\, z_{i,\mathrm{PhyAct}}.
\end{aligned}
\end{equation}
To achieve interpretability, we next apply SHAP (SHapley Additive exPlanations) value to quantify feature contributions and identify nonlinear risk patterns. This allows us to examine how circadian rhythm characteristics influence depression risk beyond black-box prediction.

\subsection{Intervention-oriented Analysis}
Prediction alone does not inform how modifiable behaviors might reduce depression risk. To address this concern, we complement screening with counterfactual and heterogeneous effect analyses focusing on modifiable circadian-related behaviors. Specifically, we investigate:

\begin{itemize}
    \item \textbf{Intervention Analysis 1:} whether \emph{intentional exercise}\footnote{We distinguish \emph{intentional exercise} from \emph{occupational physical labor}. 
Intentional exercise refers to planned, leisure-time physical activities undertaken primarily for fitness or health benefits (e.g., jogging, gym workouts, swimming). 
In contrast, occupational physical labor consists of work-related physical activities performed as part of one’s job responsibilities (e.g., construction work, farming, warehouse loading). 
Although both may involve comparable physical intensity, they differ in voluntariness, context, and psychosocial load, which may lead to distinct mental health effects.} yields greater mental health benefits than \emph{occupational physical labor} at comparable intensity levels; and
    \item \textbf{Intervention Analysis 2:} whether daytime napping compensates for insufficient nocturnal sleep.
\end{itemize}

    


\subsubsection{Model for Intervention Analysis 1:} To determine whether the mental health returns of physical activity depend on its purpose, we build the logistic regression model as
    
    \begin{equation}
    \begin{split}
            \text{logit}(P(Y=1)) = \alpha + \beta_1 x_{\text{MET}} + \beta_2 \mathbb{I}_{\text{Exercise}} \\+ \beta_3 (x_{\text{MET}} \times \mathbb{I}_{\text{Exercise}}) + \gamma \mathbf{x}_{\text{Covariates}}
    \end{split}
    \end{equation}
    

\noindent where $x_{\text{MET}}$ represents the \textit{physiological load} (MET-hours/week), i.e., the associated variable $x_{i,5}$ in Table \ref{app:control_vars}, and $\mathbb{I}_{\text{Exercise}}$ acts as the \textit{psychological context switch} (1=intentional exercise, 0=occupational labor). 
Regarding the coefficients, $\beta_1$ (baseline labor efficiency) quantifies the effect of physical activity performed solely as occupational labor; 
$\beta_2$ (contextual intercept) represents the fixed mental health difference independent of activity volume; 
and $\beta_3$ (the {intentionality bonus}) measures the differential benefit of exercise over labor. 
Finally, $\gamma \mathbf{x}_{\text{Covariates}}$ adjusts for demographics and health status.




\subsubsection{Model for Intervention Analysis 2:} To test the hypothesis that napping specifically \textit{rescues} individuals from the risks of sleep deprivation, we build the logistic regression model as
    \begin{equation}
    \begin{split}
        \text{logit}(P(Y=1)) = \alpha + \beta_1 \mathbb{I}_{\text{ShortSleep}} + \beta_2 x_{\text{Nap}} \\+ \beta_3 (\mathbb{I}_{\text{ShortSleep}} \times x_{\text{Nap}}) + \gamma \mathbf{x}_{\text{Covariates}}
    \end{split}
    \end{equation}

    


\noindent where $\mathbb{I}_{\text{ShortSleep}}$ is a binary indicator serving as a context switch (1 = sleep deprived, 0 = normal) and $x_{\text{Nap}}$ represents the {dosage} (nap duration in minutes), i.e., the associated variable $x_{i,3}$ in Table \ref{app:control_vars}. 
Regarding the coefficients, $\beta_1$ (sleep deprivation penalty) quantifies the baseline risk increase due to insufficient nocturnal sleep; 
$\beta_2$ (baseline nap effect) represents the napping effect for good sleepers; 
and $\beta_3$ (compensatory {rescue} effect) measures the additional benefit of napping specifically triggered by the sleep deprived context. 
Finally, controls $\gamma \mathbf{x}_{\text{Covariates}}$ adjusts for covariate factors.

\section{Experiments}

\subsection{Experimental Setup}
The dataset is split into a training set (70\%, \(n = 10{,}663\)) and an independent test set (30\%, \(n = 4{,}570\)) with stratified sampling to preserve class proportions. Model hyperparameters tuning are implemented via cross-validation on the training set, and all experimental results are obtained from the held-out test set.   Model performance is primarily assessed using ROC-AUC, complemented by PR-AUC and classification metrics to account for class imbalance. Confidence intervals are estimated via bootstrap resampling. Detailed experimental  settings  are provided in \textit{Appendix \ref{app:repro}}.


\subsection{Screening Performance}
The optimized LightGBM model achieves screening performance on the test set with a ROC-AUC of 0.825 and PR-AUC of 0.726 (shown in Table \ref{tab:perf_report}), indicating robust discrimination between individuals with and without elevated depressive risk. 

\begin{figure}[htbp]
    \centering
    \includegraphics[width=0.8\linewidth]{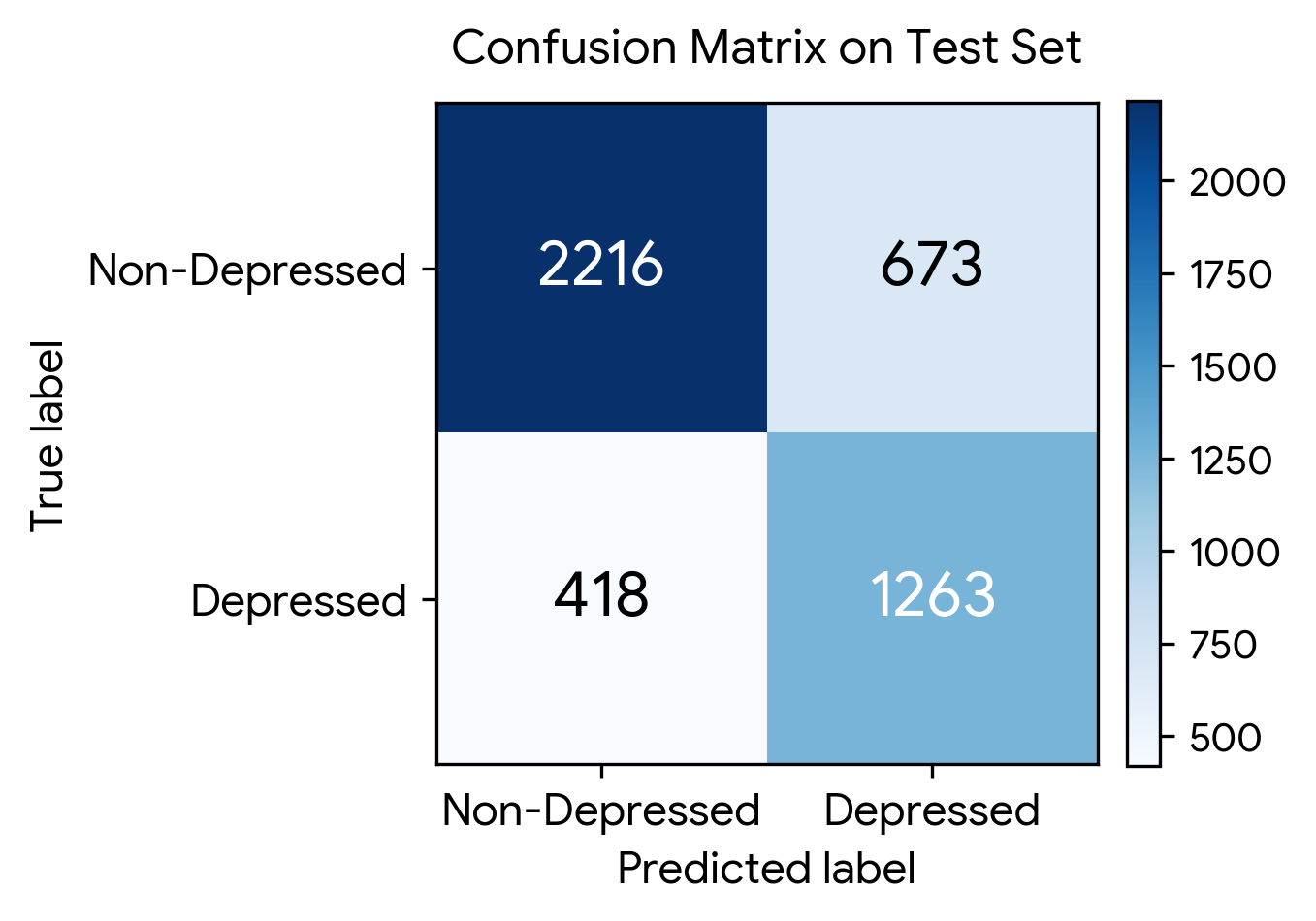}
    \Description{A 2x2 confusion matrix heatmap comparing True Labels (Y-axis) against Predicted Labels (X-axis). The classes are Non-depressed (0) and Depressed (1). The matrix shows 2216 True Negatives (top-left), 673 False Positives (top-right), 418 False Negatives (bottom-left), and 1263 True Positives (bottom-right). The heatmap colors indicate density, with the True Negative cell being the darkest blue.}
    \caption{Confusion matrix on test set.}
    \label{fig:confusion_matrix}
\end{figure}

Figure \ref{fig:confusion_matrix} reports the resulting confusion matrix. In screening contexts, sensitivity (i.e., Recall) is prioritized over specificity to minimize false negatives. The model attains a recall of 0.751 for the depressed group, while maintaining acceptable specificity for non-depressed individuals. Statistical uncertainty is quantified via nonparametric bootstrapping on the independent test set (1{,}000 resamples with replacement), with 95\%  percentile confidence intervals (2.5th--97.5th CI) reported for ROC--AUC and PR--AUC.

\begin{table}[htbp]
\caption{Classification performance on test set.}
\label{tab:perf_report}
\centering
\scriptsize
\setlength{\tabcolsep}{2pt} 
\renewcommand{\arraystretch}{1.12}
\begin{tabular}{@{}p{0.28\linewidth}cccc@{}}
\toprule
\textbf{Metric / Class} & \textbf{Precision} & \textbf{Recall} & \textbf{F1-score} & \textbf{Support} \\
\midrule

\rowcolor[HTML]{F2F2F2} \multicolumn{5}{l}{\textit{Aggregated results}} \\
Macro Average & 0.747 & 0.759 & 0.750 & 4,570 \\
Weighted Average & 0.772 & 0.761 & 0.764 & 4,570 \\
\midrule

\rowcolor[HTML]{F2F2F2} \multicolumn{5}{l}{\textit{Class-specific results}} \\
Non-depressed & 0.841 & 0.767 & 0.802 & 2,889 \\
Depressed & 0.652 & \underline{\textbf{0.751}} & 0.698 & 1,681 \\
\midrule

\rowcolor[HTML]{F2F2F2} \multicolumn{5}{l}{\textit{Probabilistic results}} \\
\textbf{Metrics} & \multicolumn{2}{c}{\textbf{Score}} & \multicolumn{2}{c}{\textbf{95\% CI}} \\
ROC-AUC & \multicolumn{2}{c}{0.825} & \multicolumn{2}{c}{[0.813, 0.838]} \\
PR-AUC  & \multicolumn{2}{c}{0.726} & \multicolumn{2}{c}{[0.703, 0.751]} \\
\bottomrule
\end{tabular}
\end{table}

\textit{Ablation study. }
Table ~\ref{tab:ablation_test_single_content_preserved} reveals two complementary roles of CRS: standalone screening and circadian-feature compression. The univariate CRS-only model, which excludes both standard covariates and the five raw circadian components, achieves a ROC-AUC of 0.7856, accuracy of 0.729, and macro-F1 of 0.722. Although this performance is lower than that of the full model, the drop is expected because a one-dimensional score necessarily discards demographic, health-status, and residual component-level information. Importantly, as a single interpretable behavioral marker, CRS substantially outperforms the non-circadian covariates-only baseline, whose ROC-AUC is 0.7223. This indicates that CRS alone carries strong depression-relevant circadian information.

When standard covariates are included, the CRS-compressed model achieves a ROC-AUC of 0.8226, only 0.0028 lower than the full model using CRS, raw circadian components, and covariates. This result supports the role of CRS as a near-lossless representation of circadian-related information in multivariable screening. Therefore, CRS should not be interpreted as a complete substitute for the full screening model; rather, it provides a compact, interpretable, and discriminative rhythm-health summary that can serve as a low-burden standalone screening signal and an efficient circadian representation for downstream risk modeling.



\begin{table}[!ht]
\caption{Ablation study results on the test set to display impact of circadian awareness.}
\label{tab:ablation_test_single_content_preserved}
\centering
\scriptsize
\setlength{\tabcolsep}{2pt} 
\renewcommand{\arraystretch}{1.02}

\begin{tabular}{@{}p{0.52\linewidth}ccc@{}}
\toprule
\textbf{Model Configuration} & \textbf{ROC-AUC} & \textbf{Accuracy} & \textbf{Macro F1} \\
\midrule
\rowcolor[HTML]{F2F2F2} \multicolumn{4}{l}{\textit{Circadian-Aware Models}} \\
CRS + Five Circadian-related Components + Covariates & \textbf{0.8254} & \textbf{0.761} & \textbf{0.750} \\
CRS + Covariates & 0.8226 & 0.749 & 0.742 \\
Univariate  CRS Only & 0.7856 & 0.729 & 0.722 \\
Circadian-related Components & 0.8242 & 0.737 & 0.733 \\
\quad w/o Sleep Quality ($z_{i,\mathrm{Quality}}$) & 0.7456 & 0.649 & 0.647 \\
\quad w/o Social Activity ($z_{i,\mathrm{Social}}$) & 0.8246 & 0.742 & 0.736 \\
\quad w/o Night Sleep Duration ($z_{i,\mathrm{Sleep}}$) & 0.8238 & 0.738 & 0.733 \\
\quad w/o Total Activity Level ($z_{i,\mathrm{Activity}}$) & 0.8253 & 0.753 & 0.745 \\
\quad w/o Nap Duration ($z_{i,\mathrm{Nap}}$) & 0.8250 & 0.756 & 0.747 \\

\midrule
\rowcolor[HTML]{F2F2F2} \multicolumn{4}{l}{\textit{Non-Circadian Model}} \\
Covariates Only & 0.7223 & 0.608 & 0.608 \\
\bottomrule
\end{tabular}
\end{table}

\subsection{Interpretability of Circadian Risk}
\begin{figure}[!ht]
    \centering
    \includegraphics[width= 0.9\linewidth]{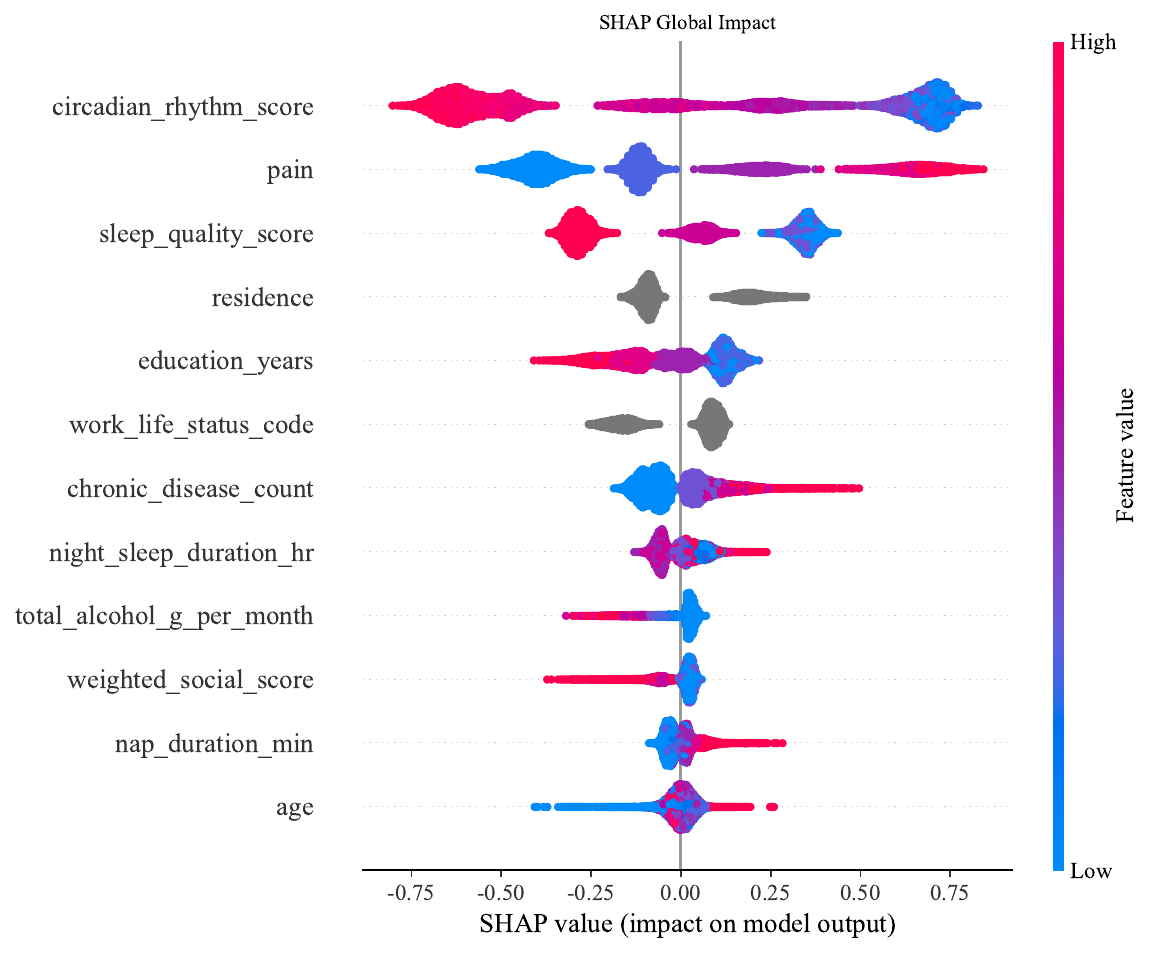}
    \Description{A SHAP summary (beeswarm) plot showing features ranked by global importance. Each point represents an individual observation. Feature values are color-coded from low (blue) to high (red). High CRS values cluster on the negative SHAP side, indicating a protective effect against depression, whereas higher pain severity and chronic disease burden shift predictions toward increased risk.}
    \caption{SHAP summary plot illustrating the global contribution of risk factors to depression risk predictions.}
    \label{fig:SHAP_Summary}
\end{figure}

\begin{figure}[!ht]
    \centering
    \includegraphics[width=\linewidth]{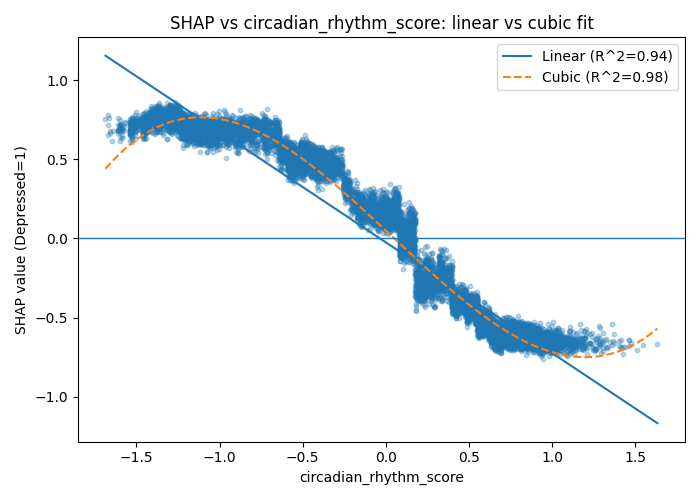}
    \Description{A scatter plot overlaying a fitted curve. The X-axis represents Standardized CRS, and the Y-axis represents SHAP values. A cubic polynomial fit (R-squared = 0.98) is shown as a red line, which decreases monotonically. The curve is flatter at the lower and higher ends of the CRS spectrum, indicating a saturation effect, and steepest in the middle range.}
    \caption{\textbf{Nonlinear association.} The cubic polynomial fit ($R^2=0.98$) effectively captures the risk saturation effect compared to the linear fit.}
    \label{fig:CRS_R}
    
\end{figure}

\textbf{SHAP analysis of risk factors.} We apply SHAP to examine how circadian rhythm characteristics influence depression risk. In Figure \ref{fig:SHAP_Summary}, each point corresponds to one individual, with color indicating the feature value from low (blue) to high (red). It presents that the global SHAP summary, where CRS emerges as the most influential predictor, surpassing individual sleep or activity variables. Notably, higher CRS values are associated with negative SHAP values, suggesting a protective effect, while severe pain and greater chronic disease burden contribute positively to predicted depression risk.  Figure ~\ref{fig:CRS_R} further visualizes the relationship between CRS (i.e., the \texttt{circadian\_rhythm\_score} in Figure ~\ref{fig:SHAP_Summary}) and its SHAP values.
The plot reveals a monotonically decreasing, S-shaped pattern with saturation at higher CRS levels, indicating diminishing marginal protective effects once circadian rhythm robustness exceeds a certain threshold.

\textbf{Logistic regression analysis of CRS.} 
We conduct a multivariable logistic regression to examine the association between CRS and depression risk, controlling for potential confounders including age, gender, education, marital status, chronic disease burden, smoking, and drinking status. 
As shown in Table~\ref{tab:logistic_regression_crs}, CRS exhibits a strong and statistically significant negative association with depression risk (\(\text{Coef}=-0.9407\), \(p<0.001\)). 

\begin{table}[H]
    \centering
\caption{Logistic regression results for the association between CRS and depression risk.}
    \label{tab:logistic_regression_crs}
    \resizebox{\columnwidth}{!}{%
    \begin{tabular}{lcccccc}
        \toprule
        \textbf{Variable} & \textbf{Coef.} & \textbf{Std. Err.} & \textbf{z} & \textbf{P$>|z|$} & \textbf{[0.025} & \textbf{0.975]} \\
        \midrule
        const & -0.2745 & 0.175 & -1.568 & 0.117 & -0.618 & 0.069 \\
        CRS\_Scaled & \textbf{-0.9407} & 0.022 & -43.300 & \textbf{0.000} & -0.983 & -0.898 \\
        age & -0.0139 & 0.002 & -6.522 & 0.000 & -0.018 & -0.010 \\
        gender & 0.2956 & 0.050 & 5.938 & 0.000 & 0.198 & 0.393 \\
        education\_years & -0.0463 & 0.005 & -9.175 & 0.000 & -0.056 & -0.036 \\
        marry & 0.0882 & 0.015 & 5.938 & 0.000 & 0.059 & 0.117 \\
        chronic\_disease\_count & 0.2526 & 0.018 & 13.906 & 0.000 & 0.217 & 0.288 \\
        smoking\_status & 0.0184 & 0.028 & 0.658 & 0.511 & -0.036 & 0.073 \\
        drinking\_status & -0.0363 & 0.016 & -2.216 & 0.027 & -0.068 & -0.004 \\
        \bottomrule
    \end{tabular}}
\end{table}

In terms of effect size, a one–standard deviation increase in CRS corresponds to an approximately \(61\%\) (\(1-0.39\))  reduction in the \emph{odds ratio} (OR) of elevated depression risk (\(\text{OR}=e^{-0.9407}\approx0.39\)). 
This finding suggests that individuals with more robust circadian rhythms are substantially less likely to exhibit depression risks.

\begin{figure}[!ht]
    \centering
    \includegraphics[width=\linewidth]{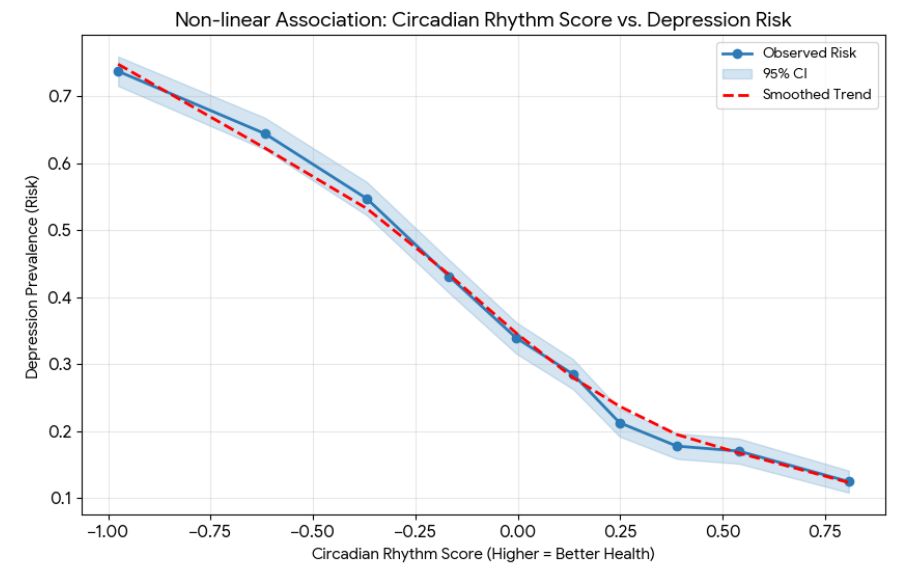}
    \Description{A line graph with a shaded confidence interval band showing the relationship between CRS (X-axis) and Depression Risk (Y-axis). The curve shows a steep, monotonic decline in depression risk as the CRS score increases from -0.5 to 0.5. The steepest reduction in risk occurs in the CRS range between 0 and 0.25.}
\caption{Observed depression risk versus normalized CRS. Higher CRS values are associated with a monotonic reduction in depression risk, with the steepest decline occurring around normalized CRS values from 0 to 0.25. Shaded ribbon indicates the 95\% CI.}
    \label{fig:CRS_Causal}
\end{figure}

Figure~\ref{fig:CRS_Causal} further illustrates a consistent monotonic relationship, whereby higher CRS values are associated with lower predicted depression risk, supporting the interpretation of CRS as a protective behavioral marker.

\begin{figure}[htbp]
    \centering
    \includegraphics[width=0.9\linewidth]{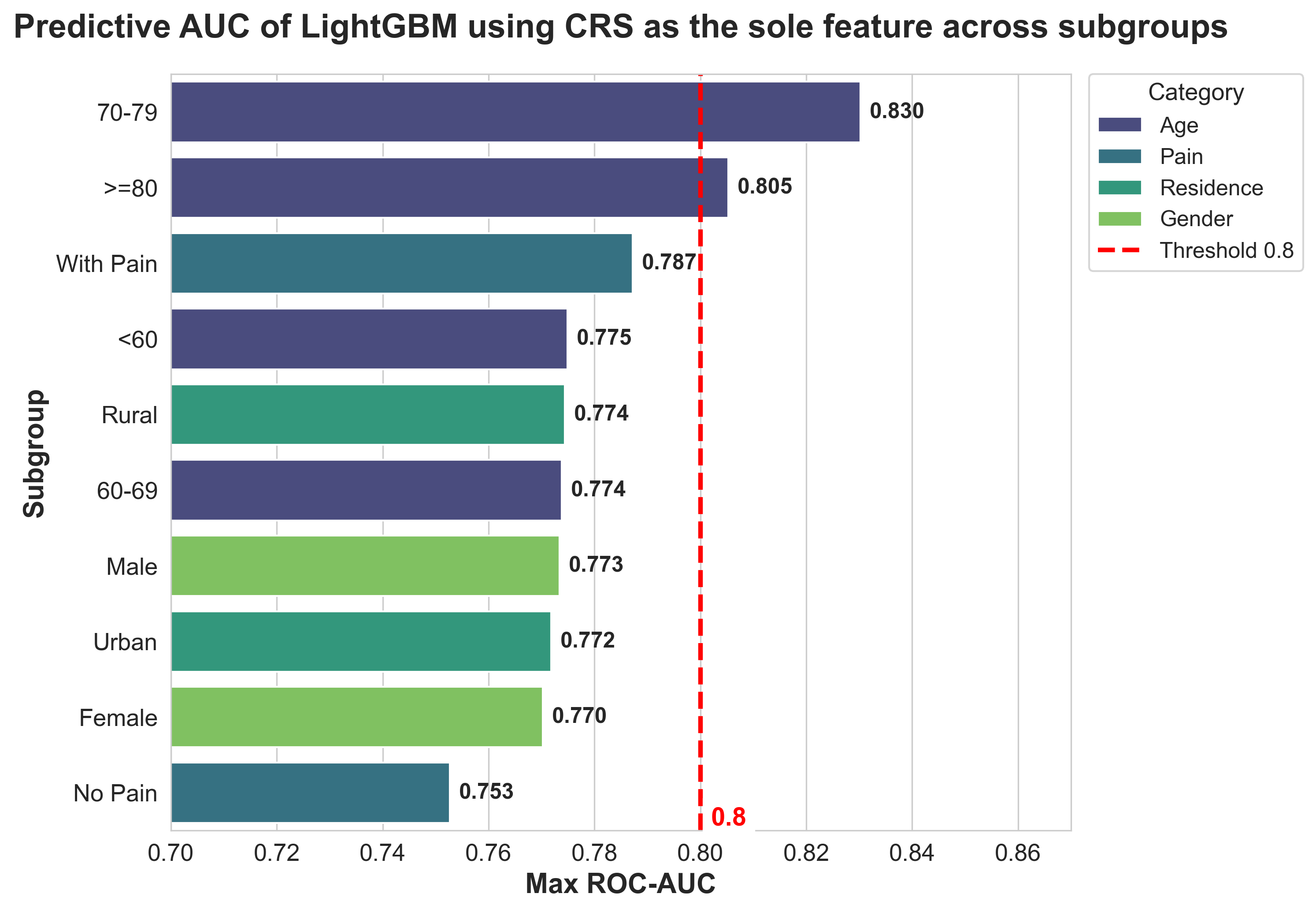}
    \Description{A bar chart or dot plot illustrating the Area Under the Curve (AUC) performance of the CRS model for different population subgroups. The AUC values are consistently high, ranging primarily between 0.80 and 0.84. The highest performance is observed in the 70-79 age group.}
    \caption{Best observed  AUC of the univariate CRS across demographic subgroups.}
    \label{fig:univariate_subgroup_auc}
\end{figure}

\textbf{Performance of univariate CRS across demographic subgroups.} We further assess the predictive performance of the CRS across diverse demographic subgroups.  As shown in Figure~\ref{fig:univariate_subgroup_auc}, univariate CRS maintains especially strong discriminative performance in older age cohorts (70--79 and 80+), achieving AUCs of 0.830 and 0.805, respectively. 

\subsection{Intervention-Oriented Experimental Findings}
Beyond risk screening, we examine how the proposed framework can provide intervention-oriented insights into modifiable behaviors.

\textbf{Answer to Intervention Analysis 1:}\textit{\textbf{Intentional exercise confers greater mental health benefits than occupational physical labor at comparable intensity levels.}}  
This conclusion is supported by a formal statistical hypothesis test using interaction-term logistic regression.

Occupational physical labor alone is not significantly associated with reduced depression risk (\(p=0.435\)) (details shown in  Table~\ref{tab:logistic_regression_interaction} of Appendix \ref{app:stats}). In contrast, the interaction between intentional exercise and activity intensity is statistically significant (\(\text{Coef.}=-0.1404\), \(P=0.006\)). This indicates that, conditional on comparable energy expenditure, engaging in intentional exercise is associated with an approximately {13.1\% (1-0.869)} depression reduction in the OR of elevated depressive risk relative to occupational physical labor (\(\text{OR}=e^{-0.1404}\approx0.869\); reduction computed as \(1-\text{OR}\)).

\begin{figure}[!ht]
    \centering
    \includegraphics[width=\linewidth]{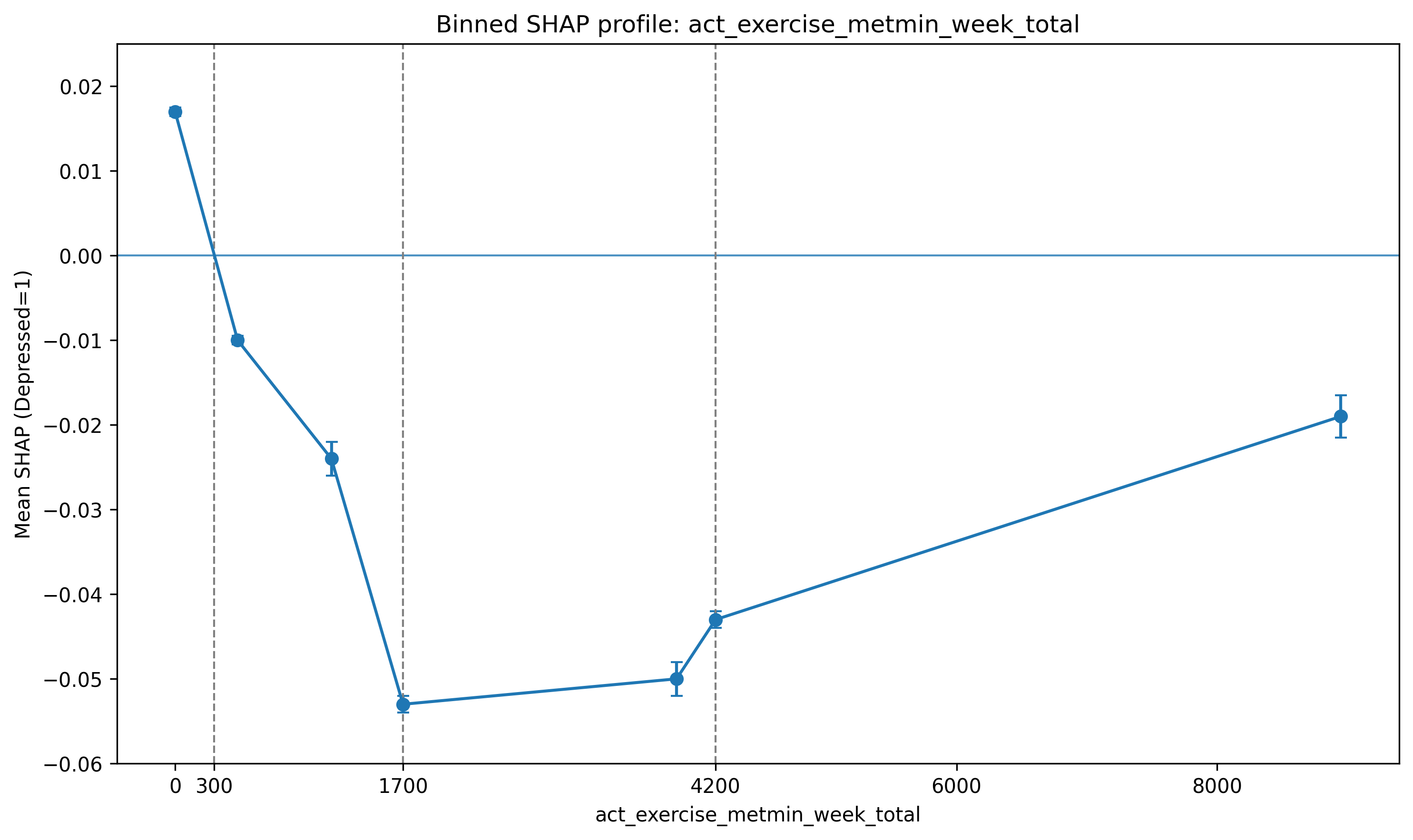}
    \Description{A SHAP dependence plot for total exercise dose (MET-min/week). The X-axis represents exercise volume, and the Y-axis represents the SHAP value. The curve crosses zero at approximately 300 MET-min/week. It shows a 'U' shape where the lowest SHAP values (greatest protection) are found in the range of 1700 to 4200 MET-min/week.}
    \caption{SHAP variation of exercise volume (MET-min/week).}
    \label{fig:exercise_total}
\end{figure}

\begin{figure}[!ht]
    \centering
    \includegraphics[width=\linewidth]{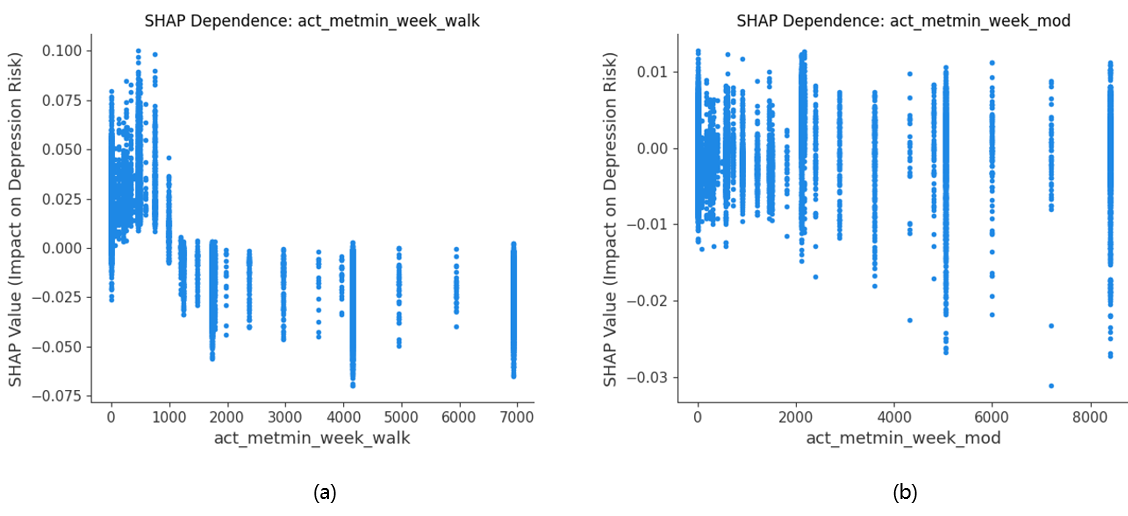}
    \Description{Two side-by-side SHAP dependence plots. Panel (a) shows 'Walking' MET-min/week, displaying a smooth curve that decreases into negative SHAP values as walking increases, indicating a stable protective effect. Panel (b) shows 'Moderate' MET-min/week, displaying a scattered cloud of points with high variance, indicating a less consistent relationship with depression risk.}
    \caption{Comparison of SHAP between (a) walking and (b) moderate intensity exercise (MET-min/week). }
    \label{fig:mod_walk_comp}
\end{figure}

\begin{figure}[!ht]
    \centering
    \includegraphics[width=\linewidth]{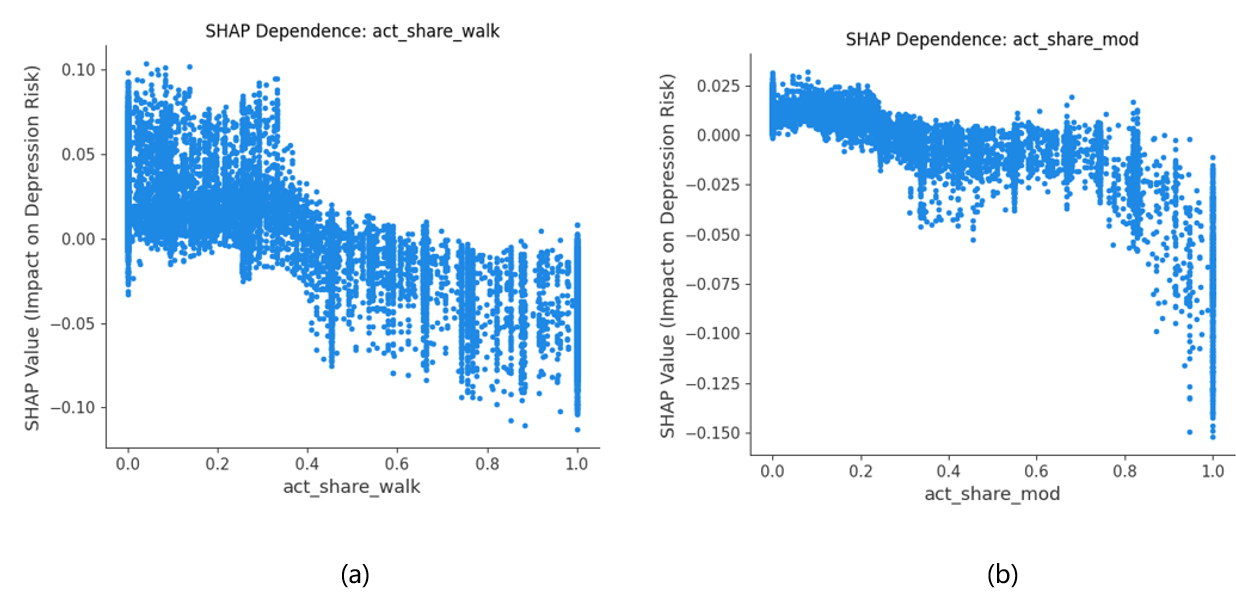}
    \Description{Two plots showing the relationship between the proportion of specific activities and SHAP values. Panel (a) shows the 'Walking Ratio', where a ratio greater than 0.4 is consistently associated with negative SHAP values (lower risk). Panel (b) shows the 'Moderate Ratio', which exhibits a noisier pattern without a clear threshold.}
    \caption{SHAP variation for (a) walking ratio and (b) moderate intensity exercise ratio in total activity.}
    \label{fig:mod_walk_share_comp}
\end{figure}

\textit{\textbf{Suggested exercise volume and activity types for older adults.}}
To inform exercise-related guidance for older adults, we analyze SHAP dependence plots for physical activity volume and conclude below suggestions:
\begin{enumerate}[label=(\arabic*)]
    \item \textit{Minimum effective dose and beneficial range.} 
    As shown in Figure~\ref{fig:exercise_total}, physical activity exhibits a non-linear protective association with depression risk. SHAP values cross zero at approximately 300~MET-min/week, indicating a minimum effective dose. The strongest protective effects are observed within the range of 1700--4200~MET-min/week, beyond which marginal benefits gradually attenuate, suggesting diminishing returns.
    
    \item \textit{Walking as a preferred activity type.} 
    Stratification by activity type reveals that walking (see Figure ~\ref{fig:mod_walk_comp} (a)) is associated with a robust and stable transition toward negative SHAP values, indicating a consistent protective association across intensity levels. In contrast, moderate intensity exercise (see Figure ~\ref{fig:mod_walk_comp} (b)) exhibits substantially higher variability, with protective effects emerging primarily at higher intensities. Moreover, composition analysis  (see Figure ~\ref{fig:mod_walk_share_comp}) suggests that maintaining walking as more than 40\% of total physical activity is stably associated with lower predicted depression risk.
\end{enumerate}

\textbf{Answer to Intervention Analysis 2:} \textit{\textbf{Daytime napping plays a compensatory role for individuals with insufficient nocturnal sleep.}} This conclusion is supported by a formal hypothesis test using interaction-term logistic regression. Interaction analysis results (detailed in Table~\ref{tab:logistic_regression_sleep_nap} of Appendix \ref{app:stats}) reveals a significant \textit{rescue effect}. While long daytime naps (\(\geq\)90 minutes) are not associated with reduced depression risk among individuals with sufficient nocturnal sleep, they are significantly protective for those with insufficient nocturnal sleep (\(<6\) hours). 
Specifically, the interaction term indicates a statistically significant reduction in depression risk (\(\text{Coef.}=-0.26\), \(P=0.03\)), corresponding to an approximately {22.9\%} reduction in the OR of elevated depressive risk (\(\text{OR}=0.771\)) for sleep deprived individuals.



\textit{\textbf{Suggested optimal napping duration (a.k.a., dose--response analysis).} }
To determine the {sweet spot} for napping intervention, we employed a counterfactual analysis based on gradient boosted decision trees. 
Unlike linear regression, this tree model captures non-linear interactions between behavioral dosage and risk response. 
We trained a tree model on the sleep deprived cohort and derived the dose-response trajectory by simulating hypothetical outcomes: we predicted depression probabilities across a continuous range of nap durations ($t \in [0, 180]$ minutes) while holding other physiological and sociodemographic covariates constant at their population central tendencies. 

\begin{figure}[!ht]
    \centering
    \includegraphics[width=\linewidth]{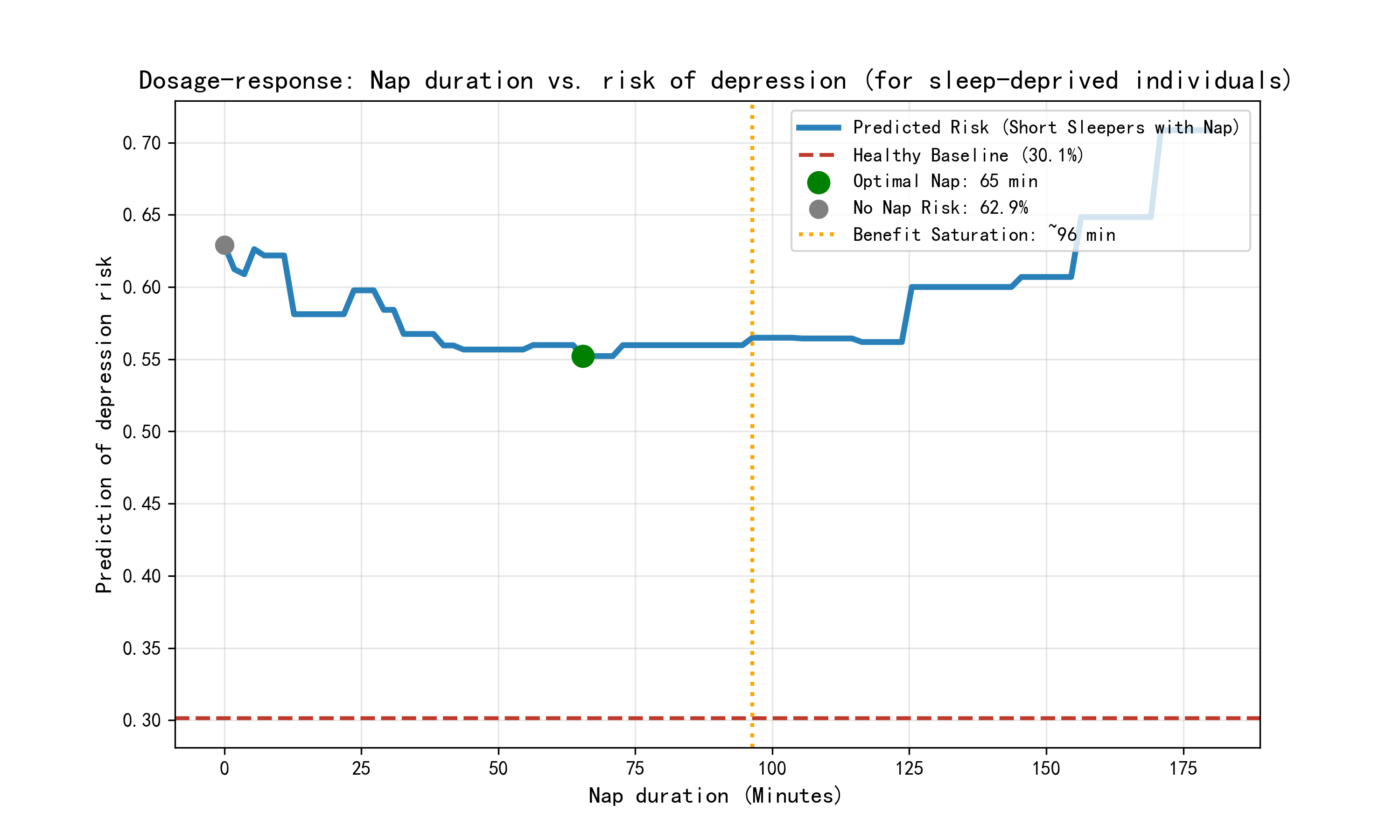}
    \Description{A dose-response curve generated by an S-Learner showing the relationship between nap duration (X-axis) and depression risk (Y-axis) for sleep deprived individuals. The curve descends rapidly from 0 minutes, reaching a minimum risk point (optimal dosage) at approximately 65 minutes, and then begins to rise or plateau after about 96 minutes.}
    \caption{\textbf{Dose-response analysis of nap duration.} For night sleep deprived individuals, the optimal nap duration is identified at $\sim$65 min, saturating beyond $\sim$96 min.}
    \label{fig:S-Learner}
\end{figure}

The resulting curve (see Figure ~\ref{fig:S-Learner}) reveals a \textit{U-shaped trajectory}. The risk of depression initially decreases with nap duration, reaching its global minimum at approximately {65 minutes}. This suggests a \textit{restorative window} where the benefits of sleep compensation peak. However, beyond this threshold---particularly exceeding 90 minutes---the risk rebounds significantly, potentially reflecting the adverse effects of sleep inertia and disruption.

\subsection{Limitations and Ethical Considerations}
\label{sec:limitations_ethics}
\textbf{Limitations.}
This study has several limitations that warrant consideration.

\textit{First, the analysis is based on cross-sectional survey data.}
Although interaction modeling and counterfactual inference technique are used to explore heterogeneous and dose-dependent associations, the results reflect associations at a single time point and do not establish long-term temporal causality. Longitudinal validation using follow-up data is needed to assess dynamic circadian changes and depression trajectories.

\textit{Second, circadian-related behaviors are self-reported.}
Measures of sleep, napping, physical activity, and social engagement may be subject to recall and reporting bias. Future studies incorporating objective behavioral measurements could further improve measurement precision and external validity.

\textit{Third, intervention-oriented analyses rely on standard causal assumptions.}
Dose--response and heterogeneous-effect estimates assume adequate control of confounding, and thus should be interpreted as decision-support and hypothesis-generating evidence, rather than definitive causal effects.

Overall, these limitations delineate the scope of inference rather than undermine the proposed framework, which is designed for scalable, interpretable depression screening and intervention-oriented data mining.


\textbf{Ethical Considerations.} This research involves the secondary analysis of de-identified data from the China Health and Retirement Longitudinal Study (CHARLS), which is publicly available for academic research. The original CHARLS data collection was approved by the Biomedical Ethics Review Committee of Peking University (IRB00001052-11015), and informed consent was obtained from all participants at the time of data collection. As this study utilizes anonymized data without direct interaction with human subjects, it is exempt from additional institutional review board (IRB) approval. Furthermore, the proposed CRS framework and depression screening models are designed as decision support tools to identify at-risk populations and generate intervention hypotheses. They are not intended to replace professional clinical diagnosis. We have explicitly analyzed the model's limitations and false-positive rates to prevent potential over-reliance on automated screening results.
\section{Conclusion and Future Work}

This study proposes a circadian rhythm-aware framework for depression screening and intervention-oriented analysis using large-scale population survey data. By integrating a supervised CRS with interpretable machine learning and causal effect estimation, the framework enables scalable identification of depression risk while preserving clinically meaningful behavioral structure. Beyond risk screening, the proposed approach provides interpretable and subgroup-specific insights into how modifiable circadian-related behaviors—such as sleep, physical activity, and daytime napping—are associated with depression risk under different circadian contexts. 

From an AI for healthcare perspective, this work illustrates how observational health data can be leveraged to bridge population-level screening, interpretability, and intervention-oriented reasoning within a unified analytical pipeline. Future work will focus on longitudinal validation using multi-wave data, integration of objective behavioral measurements, and prospective evaluation of circadian rhythm-aware decision support tools for mental health prevention.

\section*{GenAI Disclosure}
A generative AI tool \textit{Gemini3 Nanobanana Pro} was used solely for layout and formatting refinement of Figure~\ref{fig:visual_summary}. No scientific content was generated by the tool.

\begin{acks}
 This work was supported by the National Natural Science Foundation of China (Nos. 62402463, 62506049 and U2468207), the Young Scientists Fund of Shandong Province (No. ZR2023QH026), Prevention and Control of Emerging and Major Infectious Diseases-National Science and Technology Major Project (No. 2025ZD01902800), Sichuan Science and Technology Program (No. 2024NSFTD0036), and the Fundamental Research Funds for the Central Universities (No. 2682026ZT007).
\end{acks}
\clearpage

\bibliographystyle{ACM-Reference-Format}
\bibliography{reference}

\clearpage          
\appendix           
\appendix
\setcounter{table}{0}
\setcounter{figure}{0}
\renewcommand{\thetable}{A\arabic{table}}
\renewcommand{\thefigure}{A\arabic{figure}}

\section{Supplement}
\subsection{Data and Variables Details} \label{app:data}

\subsubsection{Data Exclusion Criteria}
\begin{figure}[h]
  \centering
  \includegraphics[width=0.8\linewidth]{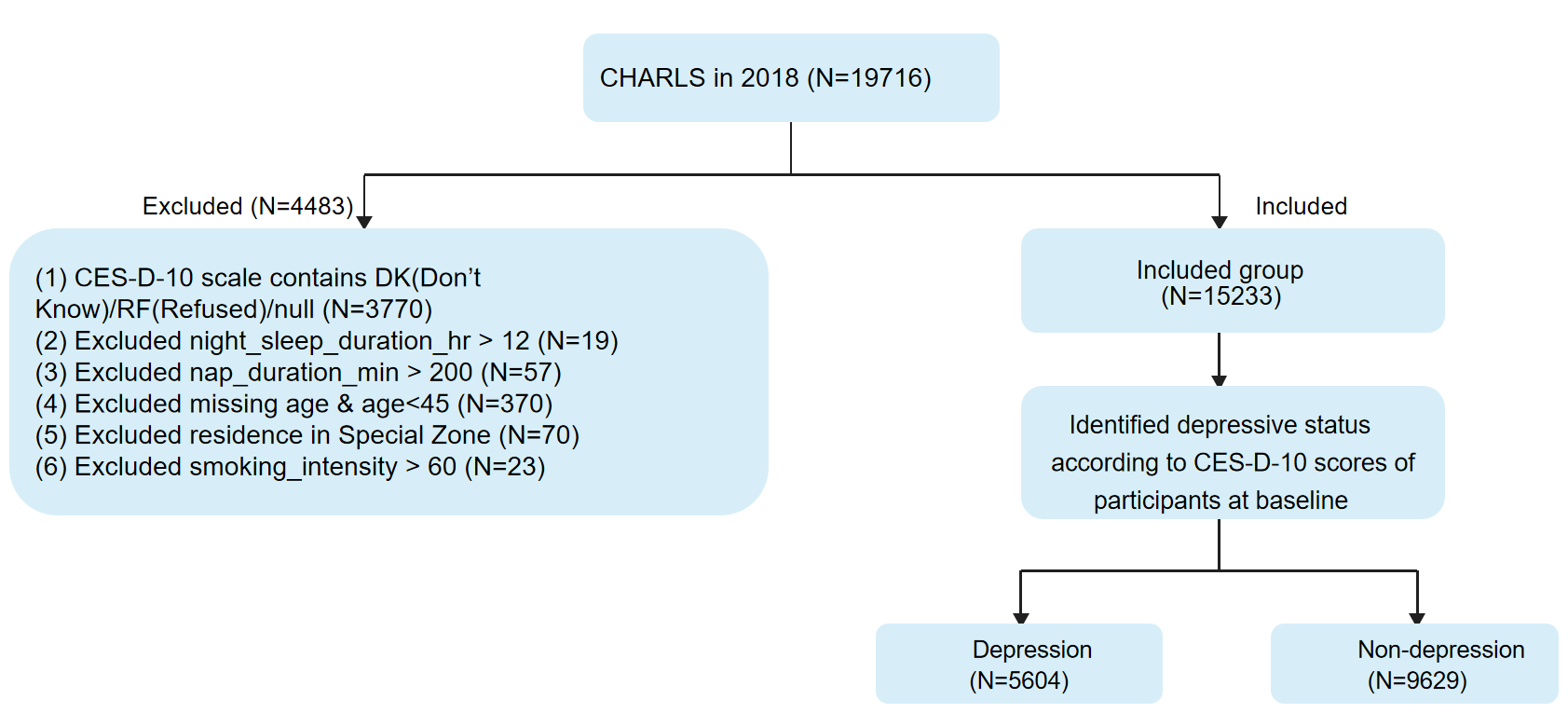}
  \Description{A flowchart diagram describing the data cleaning and participant selection process. It starts with the initial CHARLS 2018 cohort of 19,717 participants. Arrows point downwards through exclusion boxes: 'Missing Data' removes 3,770 + 370 participants; 'Outliers' removes 19 + 57 + 197 participants; 'Special Zones' removes 70 participants. The final box at the bottom shows the Analytic Sample with N equals 15,233.}
  \caption{Participant selection flowchart showing the step-by-step exclusion process.}
  \label{fig:flowchart_app}
  \vspace{-3mm}
\end{figure}

To ensure data validity, we applied the following exclusion rules to the initial CHARLS 2018 cohort ($N=19,716$):
\begin{enumerate}
    \item \textit{Missing Data:} Participants with missing responses on CES-D-10 items ($n=3,770$) or age ($n=370$) were removed.
    \item \textit{Outliers:} We excluded physically implausible values, including night sleep $>12$ hours ($n=19$), nap duration $>200$ minutes ($n=57$), and extreme smoking/alcohol values ($n=197$).
    \item \textit{Others:} Residents in special zones were excluded ($n=70$).
\end{enumerate}
The participant selection process is visualized in Figure~\ref{fig:flowchart_app}.

\subsubsection{Ground Truth}\label{app:gt}
Depressive symptoms were assessed using the 10-item Center for Epidemiologic Studies Depression Scale (CES-D-10) \cite{radloff1977ces}.  Participants rated the frequency of specific feelings during the past week on a 4-point scale: 0 (rarely or none of the time), 1 (some or a little of the time), 2 (occasionally or a moderate amount of the time), and 3 (most or all of the time). 
Items 5 ("I felt hopeful about the future") and 8 ("I was happy") were reverse-coded prior to summation. 
The total score $S_i$ ranges from 0 to 30. 
Consistent with established validation studies in Chinese elderly populations, a binary cutoff of $\ge 10$ was applied to identify individuals with elevated depressive risk ($y_i=1$).


\subsection{CRS Construction Details} \label{app:crs_details}
This section details the feature engineering and optimization process for the CRS.

\subsubsection{Definition of Normalized Circadian Component} \label{ref:weight}
Let $\mathbf{z}_i$ denote the z-score normalized circadian-related component vector for individual $i$. The five behavioral components are computed as follows:

\begin{itemize}
    \item \textbf{Sleep Quality:} Derived directly from the self-reported sleep quality score.
    \[ z_{i,SleepQuality} = \texttt{sleep\_quality\_score}_i \]
    
    \item \textbf{Nocturnal Sleep Deviation:} Quantifies the proximity to an ideal 8-hour sleep duration.
    \[ z_{i,NightSleep} = -\left| \texttt{night\_sleep\_duration\_hr}_i - 8 \right| \]
    
    \item \textbf{Napping Pattern:} A discrete scoring system rewarding short \textit{power naps} while penalizing excessive daytime sleepiness.
    \[
    z_{i,Nap} =
    \begin{cases}
    0,  & \texttt{nap\_duration\_min}_i = 0, \\
    1,  & 0 < \texttt{nap\_duration\_min}_i \le 30, \\
    -1, & \texttt{nap\_duration\_min}_i > 30.
    \end{cases}
    \]

    \item \textbf{Social Activity and Physical Activity:} 
Measured by the weighted social engagement score and total MET-hours/week, respectively. 
While the relationship between activity levels and health may be complex at extremes, we model them as monotonically positive factors in this screening context, assuming that higher engagement generally reflects better functional status in older adults.
\[
    z_{i,SocialAct} =
    \begin{cases}
    \texttt{weighted\_social\_score}_i, & \text{if available}, \\
    \texttt{social\_activity\_count}_i, & \text{otherwise}.
    \end{cases}
\]

\[ z_{i,PhyAct} = \texttt{total\_activity\_level\_met}_i \]
\end{itemize} 

\subsection{Reproducibility} \label{app:repro}


\subsubsection{Hyperparameter Tuning}
We utilized the python package \textit{Optuna} \cite{akiba2019optuna} to optimize the hyperparameters of LightGBM. 
To ensure robustness against class imbalance and overfitting, the objective function maximized the ROC-AUC score using \textit{stratified 5-fold cross-validation} with a fixed random seed.
The optimization process consisted of 100 trials.
Fixed model parameters included \texttt{objective='binary'}, \texttt{metric='auc'}, and notably \texttt{is\_unbalance=True} to automatically weight the positive class. Early stopping with a patience of 100 rounds was applied during training.
The search space of each hyperparameter and the final selected value are as follows.
\begin{itemize}
    \item \texttt{n\_estimators}: Integer search $[200, 2000]$ with step 100 (Selected: 2000)
    \item \texttt{learning\_rate}: Log-uniform search $[0.01, 0.1]$ (Selected: 0.0685)
    \item \texttt{num\_leaves}: Integer search $[20, 100]$ (Selected: 35)
    \item \texttt{max\_depth}: Integer search $[5, 20]$ (Selected: 8)
    \item \texttt{subsample}: Uniform search $[0.6, 1.0]$ (Selected: 0.9392)
    \item \texttt{colsample\_bytree}: Uniform search $[0.5, 1.0]$ (Selected: 0.5665)
    \item \texttt{reg\_alpha} (L1): Log-uniform search $[0.1, 10.0]$ (Selected: 9.8709)
    \item \texttt{reg\_lambda} (L2): Log-uniform search $[0.1, 10.0]$ (Selected: 7.9652)
\end{itemize}

\subsubsection{Experimental Implementation Details} \label{app:Experiment}

\textit{Operating Point Selection.} To avoid test-set leakage, the operating threshold $t^{*}$ was selected solely using out-of-fold (OOF) predictions from the training set. We chose the threshold that maximized the F1 score on the OOF precision--recall curve and subsequently applied this fixed threshold to the independent test set for all reported classification metrics.

\textit{Uncertainty Estimation.} Statistical uncertainty is quantified via nonparametric bootstrapping on the independent test set (1{,}000 resamples with replacement), with 95\% percentile confidence intervals (2.5th--97.5th percentiles) reported for ROC--AUC and PR--AUC.

\subsection{Detailed Statistical Results} \label{app:stats}
\paragraph{Variable coding for the activity-purpose interaction model.}
For Intervention Analysis 1, the terminology used in the main text is mapped to the regression variables in Table~\ref{tab:logistic_regression_interaction}. Let \(X_{\mathrm{MET}}\) denote the physical activity volume measured in MET-hours/week. The variables \texttt{purpose\_labor}, \texttt{purpose\_ent}, and \texttt{purpose\_exe} denote indicator variables for occupational physical labor, entertainment-related activity, and intentional exercise, respectively. The interaction variables \texttt{inter\_mets\_ent} and \texttt{inter\_mets\_exe} are constructed by multiplying \(X_{\mathrm{MET}}\) by the corresponding activity-purpose indicators. Therefore, \texttt{purpose\_labor} corresponds to the main-text statement that occupational physical labor alone is not significantly associated with reduced depression risk, whereas \texttt{inter\_mets\_exe} corresponds to the reported interaction between intentional exercise and activity intensity.

\begin{table}[!b]
    \centering
    \caption{Logistic regression results: interaction ffects of physical activity purpose on depression risk}
    \label{tab:logistic_regression_interaction}
    \resizebox{\columnwidth}{!}{%
    \begin{tabular}{lcccccc}
        \toprule
        \textbf{Variable} & \textbf{Coef.} & \textbf{Std. Err.} & \textbf{z} & \textbf{P$>|z|$} & \textbf{[0.025} & \textbf{0.975]} \\
        \midrule
        const & -0.5443 & 0.206 & -2.636 & 0.008 & -0.949 & -0.140 \\
        purpose\_labor & \textbf{-0.0217} & 0.028 & -0.781 & \textbf{0.435} & -0.076 & 0.033 \\
        purpose\_ent & -0.2177 & 0.096 & -2.279 & 0.023 & -0.405 & -0.031 \\
        purpose\_exe & -0.1539 & 0.049 & -3.132 & 0.002 & -0.250 & -0.058 \\
        inter\_mets\_ent & -0.1321 & 0.106 & -1.246 & 0.213 & -0.340 & 0.076 \\
        inter\_mets\_exe & \textbf{-0.1404} & 0.051 & -2.744 & \textbf{0.006} & -0.241 & -0.040 \\
        age & -0.0052 & 0.003 & -2.070 & 0.038 & -0.010 & -0.000 \\
        gender & 0.3305 & 0.059 & 5.604 & 0.000 & 0.215 & 0.446 \\
        education\_years & -0.0869 & 0.006 & -14.880 & 0.000 & -0.098 & -0.075 \\
        marry & 0.0910 & 0.017 & 5.298 & 0.000 & 0.057 & 0.125 \\
        chronic\_disease\_count & 0.2827 & 0.021 & 13.449 & 0.000 & 0.242 & 0.324 \\
        smoking\_status & 0.0288 & 0.033 & 0.861 & 0.389 & -0.037 & 0.094 \\
        drinking\_status & -0.0608 & 0.019 & -3.142 & 0.002 & -0.099 & -0.023 \\
        \bottomrule
    \end{tabular}}
\end{table}

\paragraph{Variable coding for the sleep--nap interaction model.}
For Intervention Analysis 2, the terminology used in the main text is mapped to the regression variables in Table~\ref{tab:logistic_regression_sleep_nap}. \textit{Insufficient nocturnal sleep} refers to participants with night sleep duration \(<6\) hours and is encoded by the binary variable \texttt{sleep\_deprived}. Daytime napping is discretized into three categories: \texttt{nap\_1\_Power} for short power naps, \texttt{nap\_2\_Restorative} for intermediate restorative naps, and \texttt{nap\_3\_Long} for long daytime naps with duration \(\geq 90\) minutes. Accordingly, \texttt{inter\_sleep\_nap\_3\_Long} represents the interaction between insufficient nocturnal sleep and long daytime napping, testing whether long naps have an additional association with depression risk specifically among individuals sleeping less than 6 hours per night. The non-significant coefficient of \texttt{nap\_3\_Long} corresponds to the main-text statement that long daytime naps are not associated with reduced depression risk among individuals with sufficient nocturnal sleep, whereas the significantly negative coefficient of \texttt{inter\_sleep\_nap\_3\_Long} supports the reported compensatory or rescue effect.

\begin{table}[!b]
    \centering
    \caption{Logistic regression results: interaction effects of napping and sleep deprivation on depression risk}
    \label{tab:logistic_regression_sleep_nap}
    \resizebox{\columnwidth}{!}{%
    \begin{tabular}{lcccccc}
        \toprule
        \textbf{Variable} & \textbf{Coef.} & \textbf{Std. Err.} & \textbf{z} & \textbf{P$>|z|$} & \textbf{[0.025} & \textbf{0.975]} \\
        \midrule
        const & -0.6434 & 0.166 & -3.879 & 0.000 & -0.968 & -0.318 \\
        sleep\_deprived & 0.9491 & 0.057 & 16.587 & 0.000 & 0.837 & 1.061 \\
        nap\_1\_Power & -0.0341 & 0.066 & -0.520 & 0.603 & -0.163 & 0.094 \\
        nap\_2\_Restorative & -0.0556 & 0.055 & -1.019 & 0.308 & -0.163 & 0.051 \\
        nap\_3\_Long & 0.0541 & 0.069 & 0.779 & 0.436 & -0.082 & 0.190 \\
        inter\_sleep\_nap\_1\_Power & -0.0022 & 0.102 & -0.021 & 0.983 & -0.202 & 0.198 \\
        inter\_sleep\_nap\_2\_Restorative & -0.0403 & 0.090 & -0.447 & 0.655 & -0.217 & 0.136 \\
        inter\_sleep\_nap\_3\_Long & \textbf{-0.2602} & 0.120 & -2.172 & \textbf{0.030} & -0.495 & -0.025 \\
        age & -0.0074 & 0.002 & -3.698 & 0.000 & -0.011 & -0.003 \\
        gender & 0.2637 & 0.047 & 5.596 & 0.000 & 0.171 & 0.356 \\
        education\_years & -0.0775 & 0.005 & -16.347 & 0.000 & -0.087 & -0.068 \\
        marry & 0.0893 & 0.014 & 6.413 & 0.000 & 0.062 & 0.117 \\
        chronic\_disease\_count & 0.2686 & 0.017 & 15.607 & 0.000 & 0.235 & 0.302 \\
        smoking\_status & 0.0142 & 0.027 & 0.534 & 0.593 & -0.038 & 0.066 \\
        drinking\_status & -0.0657 & 0.016 & -4.228 & 0.000 & -0.096 & -0.035 \\
        \bottomrule
    \end{tabular}}
\end{table}

\end{document}